\pdfoutput=1

\documentclass[11pt]{article}

\usepackage[preprint]{acl}

\usepackage{times}
\usepackage{latexsym}

\usepackage[T1]{fontenc}

\usepackage[utf8]{inputenc}

\usepackage{microtype}

\usepackage{inconsolata}

\usepackage{graphicx}
\usepackage{amsmath}
\usepackage{algorithm}
\usepackage{algpseudocode}
\usepackage{subcaption}
\usepackage{float}

\usepackage{natbib}
\usepackage{float}
\usepackage{mathtools}
\usepackage{booktabs}
\usepackage{pifont}
\definecolor{darkred}{rgb}{0.8, 0, 0}  
\definecolor{darkgreen}{rgb}{0, 0.5, 0} 
\usepackage{multirow}
\usepackage{enumitem}
\usepackage{cleveref}
\usepackage{xspace}

\newcommand{\rrrv}{\textit{R\textsuperscript{3}V}\xspace}

\title{Vision-Language Models Can Self-Improve Reasoning via Reflection}

\author{
Kanzhi Cheng\textsuperscript{1} \footnotemark[2] \quad \quad
Yantao Li\textsuperscript{1} \footnotemark[2] \quad \quad
Fangzhi Xu\textsuperscript{2} \quad \quad
Jianbing Zhang\textsuperscript{1} \\
\bf{
Hao Zhou\textsuperscript{2\,3} \quad \quad
Yang Liu\textsuperscript{2\,3} \quad \quad
}\\
\textsuperscript{1}National Key Laboratory for Novel Software Technology, Nanjing University \\
\textsuperscript{2}Shanghai AI Lab
\textsuperscript{3}Institute for AI Industry Research (AIR), Tsinghua University \\
\texttt{\{chengkz,li\textunderscore yantao\}@smail.nju.edu.cn} \quad \texttt{fangzhixu98@gmail.com} \\
\texttt{zjb@nju.edu.cn} \quad \texttt{zhouhao@air.tsinghua.edu.cn}  \quad \texttt{liuyang2011@tsinghua.edu.cn} \\
}

\begin{document}
\maketitle
\footnotetext[2]{Equal contribution.}
\footnotetext[1]{\begin{minipage}[t]{0.9\linewidth} 
Our code is available at \url{https://github.com/njucckevin/MM-Self-Improve}. 
\end{minipage}}
\begin{abstract}

Chain-of-thought (CoT) has proven to improve the reasoning capability of large language models (LLMs).
However, due to the complexity of multimodal scenarios and the difficulty in collecting high-quality CoT data, CoT reasoning in multimodal LLMs has been largely overlooked.
To this end, we propose a simple yet effective self-training framework, \rrrv, which iteratively enhances the model’s \textbf{V}ision-language \textbf{R}easoning by \textbf{R}eflecting on CoT \textbf{R}ationales.
Our framework consists of two interleaved parts: (1) iteratively bootstrapping positive and negative solutions for reasoning datasets, and (2) reflection on rationale for learning from mistakes.
Specifically, we introduce the self-refine and self-select losses, enabling the model to refine flawed rationale and derive the correct answer by comparing rationale candidates.
Experiments on a wide range of vision-language tasks show that \rrrv consistently improves multimodal LLM reasoning, achieving a relative improvement of 23\% to 60\% over GPT-distilled baselines.
Additionally, our approach supports self-reflection on generated solutions, further boosting performance through test-time computation. \footnotemark[1]

\end{abstract}

\section{Introduction}

Humans often rely on intuitive Chain-of-Thought (CoT) to perform complex reasoning \cite{ericsson1980verbal}. 
Previous studies have shown that this CoT capacity also emerges in Large Language Models (LLMs) \cite{wei2022chain}.
Through simple prompting or fine-tuning \cite{cobbe2021training, kojima2022large, hsieh2023distilling}, CoT enhances the reasoning performance of LLMs while providing insights into their decision-making process.
Recently, OpenAI o1 further advances reasoning by producing long internal CoT sequences, taking LLMs intelligence to a new level.

\begin{figure}[t]
\centering
\includegraphics[width=0.40\textwidth]{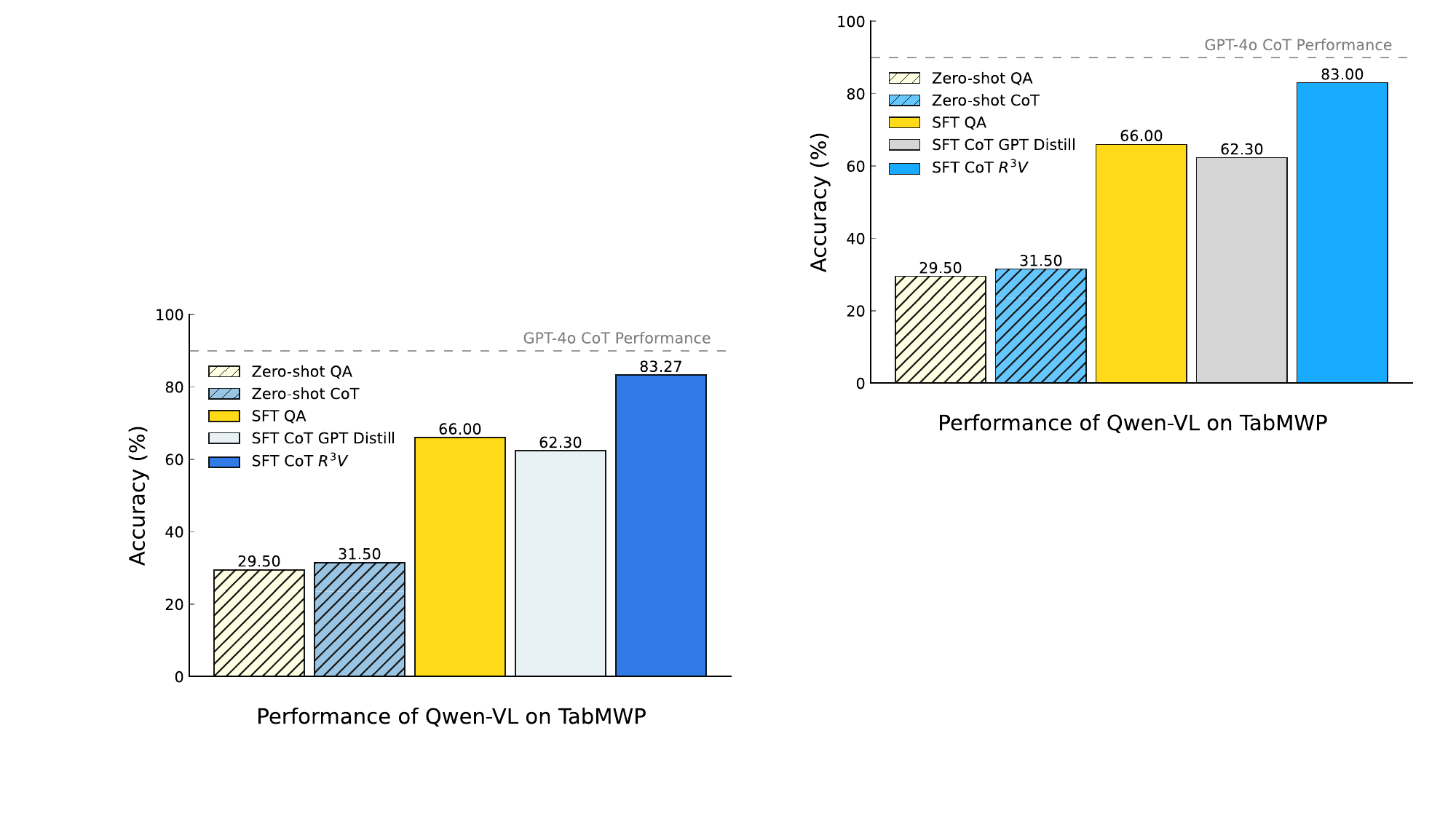}
\caption{Results of Qwen-VL on TabMWP, a visual mathematical reasoning dataset. 
Qwen-VL exhibits weak zero-shot CoT reasoning performance, while our \rrrv iteratively self-improves, surpassing the GPT-distilled baseline by a large margin.}
\label{fig:figure_1}
\vspace{-1.0em}
\end{figure}


While CoT reasoning has significantly advanced LLMs in textual domains, extending CoT to multimodal settings remains an open problem.
Unlike the abundant, unsupervised text-based CoT in pre-training corpora \cite{kojima2022large, wei2022chain}, multimodal CoT resources are scarce in the text-dominated internet collections  \cite{instructblip}, hindering the full realization of Multimodal LLMs' (MLLMs) reasoning potential.

Recent studies show that open-sourced MLLMs struggle to integrate visual cues into their reasoning process, resulting in weak CoT performance \cite{zhang2024mathverse, shi2024math}.
Consistent with our observations in \Cref{fig:figure_1}, CoT prompting provides minimal gains over direct prediction \cite{chen-etal-2024-m3cot} and falls far behind GPT-4o.
One potential solution is to construct multimodal CoT annotations for post-training; however, manual annotation is prohibitively expensive and hard to scale.
This raises our first research question: 
\emph{can MLLMs self-improve the reasoning capabilities through bootstrapping on CoT samples?}
Orthogonal to fine-tuning on curated CoT annotations, relying solely on positive samples can lead to suboptimal policy due to insufficient exploration of reasoning paths.
Inspired by human thinking, another promising direction involves learning from trial-and-errors \cite{yuan2024self, song2024trial},
where mistakes are not failures but key opportunities to enhance reasoning.
A few multimodal approaches use corrupted prompts to create negative samples for preference learning, aiming to improve image comprehension \cite{wang2023evaluation, deng2024enhancing}.
However, these methods fail to generate reasoning-aligned positive and negative CoT solutions, making them unsuitable for complex multimodal reasoning tasks.
Thus, it remains unaddressed: \emph{how can MLLMs efficiently learn from mistakes to improve their reasoning skills?}


To address the above two questions,
this paper proposes \rrrv, a self-training framework that enables the model to \textbf{R}eflect on bootstrapped CoT \textbf{R}ationales, thereby strengthening its \textbf{V}ision-Language \textbf{R}easoning.
Firstly, we leverage MLLM's pre-existing but weak CoT ability to bootstrap both rationales and answers for a given question, enabling the collection of a large number of positive and negative solutions based on answer correctness.
Secondly, we introduce a reflection mechanism on negative solutions to help the model learn from mistakes. Specifically, we design self-refine and self-select losses that guide the model to correct flawed rationales and derive the correct answer by comparing rationale candidates, respectively.
The above synergistic process can be repeated, with improved samples boosting MLLM's reasoning and the enhanced model further improving rationale generation.
Additionally, through self-select training, our model can derive the superior solution from multiple samples, further boosting performance via test-time computation.

We conduct experiments across a wide range of multimodal reasoning benchmarks, including charts, geometry, commonsense, science, mathematics, etc.
\rrrv progressively enhances the reasoning ability of MLLMs, delivering a 23\%-60\% relative accuracy improvement compared to GPT distillation,
and consistently outperforming the strong self-training baseline, STaR \cite{zelikman2022star}.
Moreover, our test-time selection is robust and effective, consistently surpassing Pass@1 and majority voting, even in OOD scenarios.

Our main contributions are as follows:
\begin{itemize}[itemsep=2pt,topsep=3pt,parsep=2pt,leftmargin=*]
    \item We introduce an iterative self-training framework \rrrv that leverages CoT bootstrapped by MLLM itself for self-improvement. To our knowledge, this is the first attempt to apply self-training in vision-language reasoning.
    \item We propose learning from mistakes through self-reflection, with support for test-time computation to further improve reasoning performance.
    \item We perform extensive evaluations across 6 different multimodal domains to validate the effectiveness of \rrrv. Our analysis reveals the key factors driving the success of multimodal self-training.
\end{itemize}

\section{Related Work}

\noindent
\textbf{Vision-Language Reasoning}
Beyond the extensively studied unimodal reasoning \cite{cobbe2021training, sun2023corex}, multimodal reasoning has recently attracted significant interest as an essential part of human intelligence \cite{yue2024mmmu, lu2023mathvista}.
Although MLLMs perform well on general vision-language benchmarks \cite{liu2024improved, chen2024far}, integrating visual cues into the reasoning process poses unique challenges, especially for open-source models \cite{zhang2024mathverse, chen-etal-2024-m3cot}.
Several studies have explored using rationale datasets to fine-tune models and enhance visual-language reasoning capabilities.
For example, \cite{gao2023g, zhang2024mavis} augmented existing mathematical datasets with rationales using GPT distillation, while \cite{yang2024mathglm} enhanced performance through manually collected CoT annotations.
In this work, we advocate for MLLMs to self-improve, reducing reliance on resource-heavy rationale annotations.

\noindent
\textbf{Self-Training Methods}
Self-training helps the model learn from its own generated outputs, reducing the need for labor-intensive human annotations \cite{yuan2024self, chen2024self}.
Prior works have focused on enhancing the reasoning capacity of LLM.
The typical approach involves sampling multiple rationales and filtering positive and negative solutions based on the answers.
The LLM is then fine-tuned on the positive samples \cite{zelikman2022star, hosseini2024v, yuan2023scaling} or improved using preference learning \cite{wang2024self, mitra2024orca}, such as DPO \cite{rafailov2024direct}.
Recent advances have also extended self-training to agents \cite{song2024trial} and neural symbolic \cite{xu2024interactive} scenarios.
In this paper, we pioneer the exploration of self-training in vision-language reasoning, investigate the failure of DPO in multimodal settings, and address these challenges with our \rrrv framework.

\section{Methodology}

\begin{figure*}[t]
\large
\centering
\includegraphics[scale=0.55]{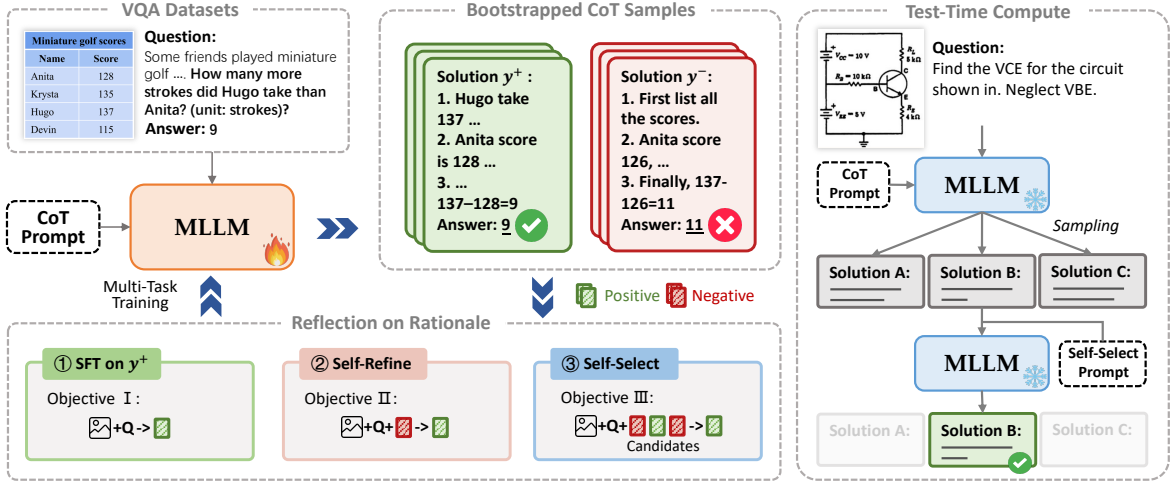}
\caption{Overview of our multimodal self-training framework of \rrrv. It boosts vision-language reasoning by iteratively reflecting on bootstrapped CoT rationales and enables self-reflection through test-time computing.}
\label{fig:frame}
\end{figure*}

Our self-training framework consists of two alternating components: 
(1) bootstrapping a large number of positive and negative CoT solutions for multimodal questions (\Cref{sec:pre});
(2) using the above-sampled solutions to reflect on rationales and learn from mistakes (\Cref{sec:r3v}).
This iterative process turns the MLLM from weak to strong.
The overall framework is illustrated in \Cref{fig:frame}.

\subsection{Preliminaries}
\label{sec:pre}

In visual-language reasoning, given an image $I$ and a question $x$, a multimodal large language model is required to integrate information from both the image and the question for reasoning, generating a CoT rationale $r$ and then deriving the final answer $a$.
However, due to the difficulty in collecting high-quality rationale data, constructing large-scale $(I, x, r, a)$ pairs presents significant challenges.
This hinders the enhancement of MLLM reasoning capacities through fine-tuning.
To overcome this limitation, we propose leveraging the MLLM's pre-existing but weak CoT capability to iteratively augment $(I, x, r, a)$ pairs from the widely available vision question answering data $(I, x, a)$, enabling the model to self-improve.

Following STaR \cite{zelikman2022star}, the MLLM self-training process involves iteratively fine-tuning on its self-generated rationale data.
In each iteration $t$, given a question $x$ from training set $\mathcal{D} = \{(I, x, \hat a)\}$, the MLLM $\mathcal{M}$ first generate a CoT rationale $r$ along with an answer $a$
, formulated as $\{(r_i, a_i)\}_{i=1}^{|\mathcal{D}|}$. These intermediate outputs are then combined with the original training set, resulting in an augmented dataset that includes rationales:
\begin{equation}
    \mathcal{D}_r = \{(I_i, x_i, r_i, a_i)\}_{i=1}^{|\mathcal{D}|}
\end{equation}
Assuming that rationales leading to correct answers are of higher quality compared to those that do not, 
we can divide $D_r$ into positive and negative sample sets based on the correctness of the answers:
\begin{equation}
    \mathcal{D}_r^{+} = \{(I_i, x_i, r_i, a_i) \mid a_i = \hat a_i\}_{i=1}^{|\mathcal{D}|}
\end{equation}
\begin{equation}
    \mathcal{D}_r^{-} = \{(I_i, x_i, r_i, a_i) \mid a_i \neq \hat a_i\}_{i=1}^{|\mathcal{D}|}
\end{equation}
We then fine-tune model $\mathcal{M}$ on the filtered positive CoT samples $\mathcal{D}_r^{+}$ using supervised fine-tuning (SFT) with a negative log-likelihood objective:
\begin{equation}
    \mathcal{L}_{SFT} = \; - \sum_{\mathclap{(I, x, y)\sim \mathcal{D}_t}} \; \log \mathcal{M}(y \mid x, I),
\end{equation}
where the $y = (r, a)$ is the solution generated by the model.
We continue repeating the above process, generating new rationales with the newly fine-tuned model, until performance plateaus.

\subsection{\texorpdfstring{\rrrv: Reflection on Rationales}{r3v: Reflection on Rationales}}
\label{sec:r3v}

The above self-improvement process strengthens the model using positive solutions, while negative ones are typically discarded.
However, negative samples comprise a large portion of the sampled solutions and offer valuable insights for further model enhancement \cite{an2023learning, hosseini2024v}.
In our preliminary experiments, we found that the noisy nature of CoT in multimodal scenarios leads to suboptimal performance when using DPO \cite{rafailov2024direct}.
Inspired by the error-driven learning of humans, we introduce reflection on rationales, teaching the model to correct its own mistakes and reflect on multiple reasoning paths to identify the correct solution.
Specifically, we propose additional self-refine (\Cref{sec:refine}) and self-select (\Cref{sec:select}) losses for multitask learning.
Our framework harnesses the continuous production of positive and negative samples in self-training, offering a robust and effective solution for learning from mistakes. \Cref{app:data_example} provides examples of different components in \rrrv.

\subsubsection{Self-Refine}
\label{sec:refine}

Upon failing to solve a problem, human students will analyze the errors in their solutions and reflect on how to correct them.
Inspired by this, we designed the self-refine mechanism to encourage the model to correct flaws in its generated solutions.
Multiple positive and negative solutions sampled during self-training can be viewed as the model's repeated reasoning on the same problem, making them well-suited for self-refine training.
Specifically, we construct dataset for self-refine as follows: 
\begin{equation}
    \mathcal{D}_{REF} = \{(I_i, x_i, y_i^{+}, y_i^{-}) \mid \exists\, y_i^{+}, y_i^{-} \}_{i=1}^{|\mathcal{D}|},
\end{equation}
where $y_i^{+}$ and  $y_i^{-}$ are positive and negative samples obtained from preceding iterations.
Next, the self-refine loss is employed to guide the model in correcting errors in its self-generated answers:
\begin{equation}
    \mathcal{L}_{REF} = \; - \sum_{\mathclap{(I, x, y^{+}, y^{-})\sim \mathcal{D}_{REF}}} \; \log \mathcal{M}(y^{+} \mid y^{-}, x, I)
\end{equation}
Throughout the self-training iterations, samples for self-refine are continuously updated to incorporate higher-quality positive solutions and harder negative solutions.

\subsubsection{Self-Select}
\label{sec:select}

Our early explorations reveal a key challenge in MLLM reasoning:
current MLLMs frequently make simple errors such as misreading chart numbers or calculation mistakes, however, the autoregressive model has no mechanism to correct them, leading to suboptimal performance.
In contrast, human reasoners implicitly simulate multiple reasoning paths, check for errors, and select the best one.
Inspired by this, we introduce the self-selection mechanism, guiding MLLMs to derive the correct answer from multiple candidate solutions.


Given a set of sampled rationales, the model is required to analyze their differences and finally select the correct answer.
Specifically, we construct the self-select dataset as follows: 
\begin{equation}
    \mathcal{D}_{SEL} = \{(I_i, x_i, \hat a_i, \mathcal{C}_i) \mid \exists\, \mathcal{C}_i \, \}_{i=1}^{|\mathcal{D}|},
\end{equation}
where $\hat a$ is the ground truth and $\mathcal{C}_i = (y_i^1, y_i^2, ..., y_i^N)$ is a set of $N$ sampled rationale-answer pair. 
In our experiments, $N$ is set to 3 by default.
We ensure that the candidate set $\mathcal{C}$ contains at least one positive solution $y^{+}$, allowing the model to select the final correct answer.
Then, the self-select loss is defined as:
\begin{equation}
    \mathcal{L}_{SEL} = \; - \sum_{\mathclap{(I, x, \hat a, \mathcal{C})\sim \mathcal{D}_{SEL}}} \; \log \mathcal{M}(\hat a|x, I, \mathcal{C})
\end{equation}

Finally, our framework combines three loss functions in a multi-task training setup to enhance MLLM reasoning (see algorithm in \Cref{app:alg}):

\begin{equation}
    \mathcal{L}_{\rrrv} = \mathcal{L}_{SFT} + \mathcal{L}_{REF} + \mathcal{L}_{SEL}
\end{equation}
From another perspective, we argue that this multi-task training enables MLLMs to learn reasoning from easy to hard: selecting the correct solution from multiple candidates, refining existing rationales, and eventually generating solutions directly.



\subsubsection{Test-Time Selection}
\label{sec:test_select}

Through self-select training, our framework enables MLLMs to reflect on their self-generated solutions and select the final answer from multiple reasoning paths.
During inference, given a question $x$ and corresponding image $I$, we first sample multiple reasoning solutions to form the candidate set $\mathcal{C}$.
Next, the MLLM is prompted to select the best answer from these candidate solutions: $a = \mathcal{M}(x, I, \mathcal{C})$.

Test-time selection offers a novel approach for MLLMs to tackle complex multimodal reasoning.
Instead of directly generating an answer, the model applies an elimination method by comparing different reasoning paths and checking for errors (e.g., visual recognition, calculation, or reasoning mistakes) to identify the most likely correct solution.
In this way, our approach further boosts reasoning performance through test-time computation.

\begin{table*}[ht]
\centering
\footnotesize
\renewcommand\arraystretch{1.1}
\resizebox{0.95\linewidth}{!}{
\begin{tabular}{lcccccccc}
\toprule
\multicolumn{1}{l|}{}                                   & \multicolumn{1}{c|}{}                                   & \multicolumn{3}{c|}{\textbf{Logical and Numerical reasoning}}                 & \multicolumn{1}{c|}{\textbf{Agentic}} & \multicolumn{1}{c|}{\textbf{Geometry}} & \multicolumn{1}{c|}{\textbf{Multi-Domain}} &                                \\
\multicolumn{1}{l|}{\multirow{-2}{*}{\textbf{Methods}}} & \multicolumn{1}{c|}{\multirow{-2}{*}{\textbf{Is CoT?}}} & \textbf{TabMWP} & \textbf{ChartQA} & \multicolumn{1}{c|}{\textbf{CLEVR-Math}} & \multicolumn{1}{c|}{\textbf{MiniWob}} & \multicolumn{1}{c|}{\textbf{GeoQA}}    & \multicolumn{1}{c|}{\textbf{M\textsuperscript{3}CoT}}        & \multirow{-2}{*}{\textbf{Avg}} \\ 
\midrule
\multicolumn{1}{l|}{GPT-4o}                             & \multicolumn{1}{c|}{\textcolor{darkgreen}{\ding{51}}}                                & 94.47           & 67.00             & \multicolumn{1}{c|}{70.60}                & \multicolumn{1}{c|}{98.50}             & \multicolumn{1}{c|}{55.17}                  & \multicolumn{1}{c|}{65.85}                      & 70.62                               \\
\midrule
\multicolumn{9}{c}{\cellcolor{gray!25}\textbf{Qwen-VL}}                                                                                                                                                                                                                                                                                                     \\ 
\midrule
\multicolumn{1}{l|}{Zero-shot QA}                       & \multicolumn{1}{c|}{\textcolor{darkred}{\ding{55}}}                                 & 29.50           & 38.56                 & \multicolumn{1}{c|}{17.32}                    & \multicolumn{1}{c|}{\textit{-}}                & \multicolumn{1}{c|}{15.14}                  & \multicolumn{1}{c|}{31.28}                      & 26.36                              \\ 
\multicolumn{1}{l|}{Zero-shot CoT}                      & \multicolumn{1}{c|}{\textcolor{darkgreen}{\ding{51}}}                                & 31.50           & 37.59                 & \multicolumn{1}{c|}{12.61}                    & \multicolumn{1}{c|}{\textit{-}}              & \multicolumn{1}{c|}{16.58}                  & \multicolumn{1}{c|}{30.73}                      & 25.80                              \\
\midrule
\multicolumn{1}{l|}{\vspace{-2pt}\textbf{SFT Based}}                 & \multicolumn{1}{c|}{}                                   &                 &                  & \multicolumn{1}{c|}{}                    & \multicolumn{1}{c|}{}                 & \multicolumn{1}{c|}{}                  & \multicolumn{1}{c|}{}                      &                                \\
\multicolumn{1}{l|}{QA}                                 & \multicolumn{1}{c|}{\textcolor{darkred}{\ding{55}}}                                 & 66.00           & 46.64            & \multicolumn{1}{c|}{65.20}               & \multicolumn{1}{c|}{\textit{-}}                & \multicolumn{1}{c|}{33.03}                  & \multicolumn{1}{c|}{48.96}                 & 51.97                               \\
\multicolumn{1}{l|}{GPT Distill}                        & \multicolumn{1}{c|}{\textcolor{darkgreen}{\ding{51}}}                                & 62.30           & 46.72            & \multicolumn{1}{c|}{51.83}               & \multicolumn{1}{c|}{51.11}            & \multicolumn{1}{c|}{31.43}             & \multicolumn{1}{c|}{47.41}                 & 48.47                             \\ 
\midrule
\multicolumn{1}{l|}{\vspace{-2pt}\textbf{Self-Train Based}}          & \multicolumn{1}{c|}{}                                   &                 &                  & \multicolumn{1}{c|}{}                    & \multicolumn{1}{c|}{}                 & \multicolumn{1}{c|}{}                  & \multicolumn{1}{c|}{}                      &                                \\
\multicolumn{1}{l|}{STaR}                               & \multicolumn{1}{c|}{\textcolor{darkgreen}{\ding{51}}}                                & 77.84           & 53.60            & \multicolumn{1}{c|}{61.45}               & \multicolumn{1}{c|}{78.22}                 & \multicolumn{1}{c|}{34.08}             & \multicolumn{1}{c|}{50.47}                 & 59.28                              \\
\multicolumn{1}{l|}{\rrrv}                                & \multicolumn{1}{c|}{\textcolor{darkgreen}{\ding{51}}}                                & \textbf{83.27}  & \textbf{57.36}   & \multicolumn{1}{c|}{\textbf{68.81}}      & \multicolumn{1}{c|}{\textbf{82.89}}   & \multicolumn{1}{c|}{\textbf{39.25}}    & \multicolumn{1}{c|}{\textbf{54.66}}        & \multicolumn{1}{l}{\textbf{64.37}}           \\ 
\midrule
\multicolumn{9}{c}{\cellcolor{gray!25}\textbf{LLaVA-1.5}}                                                                                                                                                                                                                                                                                                   \\ 
\midrule
\multicolumn{1}{l|}{Zero-shot QA}                       & \multicolumn{1}{c|}{\textcolor{darkred}{\ding{55}}}                                 & 17.66           & 13.04                 & \multicolumn{1}{c|}{19.04}                    & \multicolumn{1}{c|}{\textit{-}}                 & \multicolumn{1}{c|}{26.92}                  & \multicolumn{1}{c|}{36.63}                      & \multicolumn{1}{l}{22.66}           \\
\multicolumn{1}{l|}{Zero-shot CoT}                      & \multicolumn{1}{c|}{\textcolor{darkgreen}{\ding{51}}}                                & 15.33           & 8.39                 & \multicolumn{1}{c|}{13.87}                    & \multicolumn{1}{c|}{\textit{-}}                 & \multicolumn{1}{c|}{23.47}                  & \multicolumn{1}{c|}{35.81}                      & \multicolumn{1}{l}{19.37}           \\ 
\midrule
\multicolumn{1}{l|}{\vspace{-2pt}\textbf{SFT Based}}                 & \multicolumn{1}{c|}{}                                   &                 &                  & \multicolumn{1}{c|}{}                    & \multicolumn{1}{c|}{}                 & \multicolumn{1}{c|}{}                  & \multicolumn{1}{c|}{}                      & \multicolumn{1}{l}{}           \\
\multicolumn{1}{l|}{QA}                                 & \multicolumn{1}{c|}{\textcolor{darkred}{\ding{55}}}                                 & 48.06                & 27.20                 & \multicolumn{1}{c|}{75.08}                    & \multicolumn{1}{c|}{\textit{-}}                 & \multicolumn{1}{c|}{42.17}             & \multicolumn{1}{c|}{52.63}                 & \multicolumn{1}{l}{49.03}           \\
\multicolumn{1}{l|}{GPT Distill}                        & \multicolumn{1}{c|}{\textcolor{darkgreen}{\ding{51}}}                                & 44.63           & 28.48            & \multicolumn{1}{c|}{56.52}               & \multicolumn{1}{c|}{60.44}                 & \multicolumn{1}{c|}{33.81}             & \multicolumn{1}{c|}{47.54}                 & \multicolumn{1}{l}{45.24}           \\ 
\midrule
\multicolumn{1}{l|}{\vspace{-2pt}\textbf{Self-Train Based}}          & \multicolumn{1}{c|}{}                                   &                 &                  & \multicolumn{1}{c|}{}                    & \multicolumn{1}{c|}{}                 & \multicolumn{1}{c|}{}                  & \multicolumn{1}{c|}{}                      & \multicolumn{1}{l}{}           \\
\multicolumn{1}{l|}{STaR}                               & \multicolumn{1}{c|}{\textcolor{darkgreen}{\ding{51}}}                                & 56.67                & 33.44                 & \multicolumn{1}{c|}{73.46}                    & \multicolumn{1}{c|}{76.00}                 & \multicolumn{1}{c|}{41.25}             & \multicolumn{1}{c|}{54.06}                 & \multicolumn{1}{l}{55.81}           \\
\multicolumn{1}{l|}{\rrrv}                                & \multicolumn{1}{c|}{\textcolor{darkgreen}{\ding{51}}}                                &  \textbf{59.30}               &  \textbf{33.92}                & \multicolumn{1}{c|}{\textbf{79.01}}                    & \multicolumn{1}{c|}{\textbf{80.11}}                 & \multicolumn{1}{c|}{\textbf{45.76}}             & \multicolumn{1}{c|}{\textbf{56.08}}                      & \multicolumn{1}{l}{\textbf{59.03}}           \\ 
\bottomrule
\end{tabular}
}
\caption{Main results on six vision-language reasoning benchmarks. \textit{Is CoT?} column indicates whether a CoT or a direct answer was generated. \textit{Avg.} column reports the average performance across all tasks (\textit{-} indicates MiniWob is not applicable to this setting and is excluded from the average). \rrrv significantly improves upon the GPT-distilled baseline without additional annotation costs, and surpasses the strong baseline STaR by a large margin.}
\label{tab:main}
\end{table*}

\section{Experiments}

In our experiments, we focus on a diverse and comprehensive set of vision-language reasoning tasks to demonstrate the effectiveness of \rrrv.
We begin by outlining the benchmarks (\Cref{exp:dataset}) and experimental setup (\Cref{exp:setting}), followed by the main results of \rrrv on six widely used datasets (\Cref{exp:result}).
We also evaluated the improvements achieved by our framework in out-of-distribution (OOD) scenarios (\Cref{exp:ood}).

\subsection{Datasets}
\label{exp:dataset}

We validate our framework's self-improvement on six vision-language reasoning benchmarks, which require integrating visual information into complex, multi-step reasoning. Refer to \Cref{app:bench} for detailed information of these benchmarks.

\noindent
\textbf{TabMWP} \cite{lu2022dynamic}: A dataset for table-based math word problems requiring reasoning and numerical calculation.

\noindent
\textbf{ChartQA} \cite{masry2022chartqa}: Focuses on reasoning and calculations within real-world charts.

\noindent
\textbf{CLEVR-Math} \cite{lindstrom2022clevr}: Compositional reasoning over abstract figures.

\noindent
\textbf{MiniWob} \cite{shi2017world}: A widely-used multimodal web navigation benchmark requiring models to generate multi-step actions.

\noindent
\textbf{GeoQA} \cite{chen2021geoqa}: A geometry problem benchmark requiring complex reasoning.

\noindent
\textbf{M\textsuperscript{3}CoT} \cite{chen-etal-2024-m3cot}: A recently introduced dataset featuring multi-domain, multi-step multimodal reasoning problems.

\begin{figure*}[ht]
\centering
\includegraphics[width=0.98\textwidth]{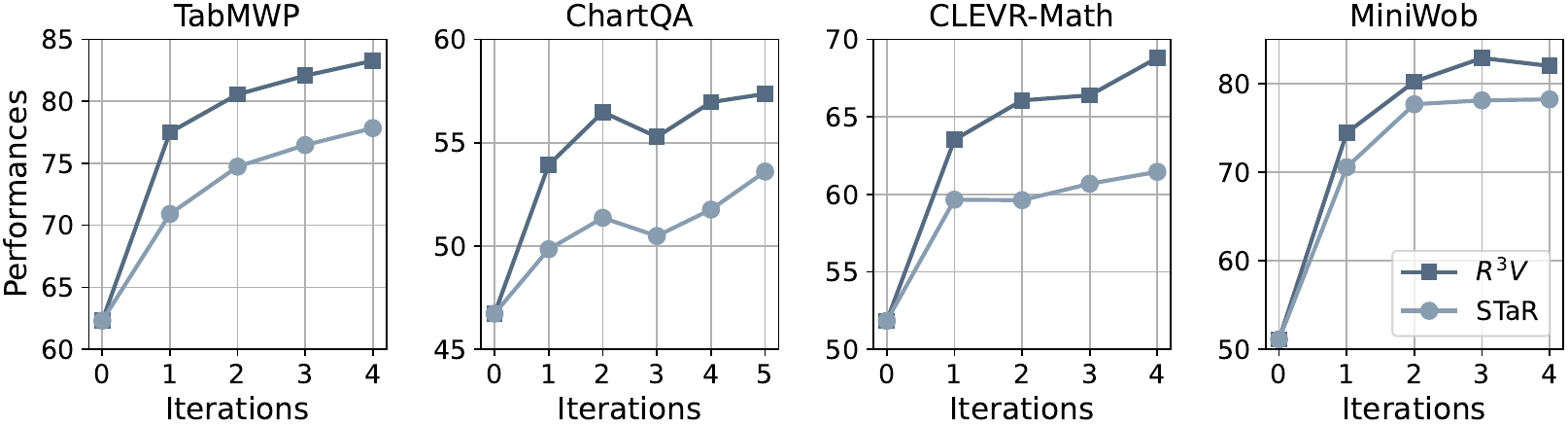}
\caption{Comparison of the iterative self-training process between \rrrv and STaR on Qwen-VL across four benchmarks. Full results are provided in \Cref{app:evo}. \rrrv demonstrates higher efficiency in evolution and superior final performance.}
\label{fig:evolution}
\vspace{-1.0em}
\end{figure*}

\subsection{Experimental Setings}
\label{exp:setting}

We primarily compare our framework with three categories of methods to comprehensively assess its effectiveness. All experiments are conducted under the same parameters to ensure a fair comparison.

\noindent
\textbf{Zero-shot Methods.}
We evaluated the MLLMs' zero-shot performance under the direct prompt (where the model tends to provide an immediate answer \cite{liu2024improved}) and the CoT prompt using "Let's think step by step." GPT-4o was also chosen as a strong baseline for comparison.

\noindent
\textbf{Supervised Fine-tuning Baselines.}
Since the self-training requires existing $(I, x, a)$ datasets, we provide the results of fine-tuning MLLMs using direct prompts on these question-answer pairs.
We also include a GPT distillation baseline, where GPT-4o annotates CoT rationales for a small subset of each dataset, and then the open-source MLLMs are fine-tuned for CoT reasoning.

\noindent
\textbf{Self-Training Methods.}
We employ the aforementioned GPT-distilled, warmed-up MLLM as the starting point for self-training, iteratively sampling positive and negative rationales from training samples for continuous self-improvement.
We then compare \rrrv with the well-known self-training baselines STaR \cite{zelikman2022star}, which iteratively fine-tunes on self-generated positive solutions for model improvement.

We use two established MLLMs, Qwen-VL \cite{bai2023qwen} and LLaVA-1.5 \cite{liu2024improved}, as base models for self-training.
We sample three solutions per sample in each iteration by default. The total number of iterations for these tasks is set to 4-5, depending on the convergence speed. Further details can be found in \Cref{app:train_detail}.

\subsection{Main Results}
\label{exp:result}

\Cref{tab:main} presents the evaluation results of \rrrv on various multimodal reasoning tasks, including logical and numerical reasoning, agentic tasks, geometry, and multi-domain scenarios. The evolution progress of self-training is illustrated in \Cref{fig:evolution}.

\paragraph{Self-training effectively converts MLLMs from weak to strong.}
Open-source MLLMs struggle with complex vision-language reasoning tasks. CoT reasoning with the "Let's think step by step" prompt (Zero-shot CoT) proves ineffective, with performance even worse than direct prompting (Zero-shot QA).
In this situation, the self-training method leverages MLLMs' pre-existing but weak CoT capabilities to bootstrap multimodal CoT data for self-improvement.
This process progressively elevates MLLMs' CoT reasoning, as shown in \Cref{fig:evolution}, taking it to the next level on top of the GPT-distilled baseline.
As an example with Qwen-VL, our self-training framework \rrrv delivers an average 32.8\% relative performance improvement over the GPT-distilled baseline (48.47 $\to$ 64.37).
This result highlights the remarkable potential of MLLMs to enhance their reasoning capabilities through self-training on synthetic data.

\paragraph{\rrrv further enhances self-training efficiency by learning from mistakes.}
Instead of discarding valuable negative samples, our \rrrv framework leverages carefully designed self-refine and self-select mechanisms to learn from negative solutions, surpassing the strong self-training baseline STaR by a large margin (average 59.28 $\to$ 64.37 on Qwen-VL).
As shown in \Cref{fig:evolution}, \rrrv demonstrates swift adaptation across different multimodal scenarios, achieving notably higher gains in the first iteration compared to the STaR baseline, highlighting the efficiency of our method.
These results underscore the value of learning from mistakes in multimodal reasoning and demonstrate the effectiveness of our reflection-based methodology.

\subsection{Out-of-Distribution Evaluation}
\label{exp:ood}

\begin{table}[t]
\centering
\footnotesize
\renewcommand\arraystretch{1.1}
\resizebox{\linewidth}{!}{
\begin{tabular}{l|c|c|c}
\toprule
\textbf{Methods}         & \textbf{MMMU}  & \textbf{MathVista} & \textbf{VCR}   \\
\midrule
Qwen-VL                  & 30.44          & 29.1               & 34.02          \\
\quad + GPT-distilled    & 33.67          & 32.7               & 45.39          \\
\quad + Ours                     & 35.63          & 35.10              & 50.23          \\
\quad + Ours (TTS) & \textbf{38.48} & \textbf{35.80}     & \textbf{51.78} \\
\bottomrule
\end{tabular}
}
\caption{Evaluation results on OOD benchmarks. \textit{Ours (TTS)} denotes Test-time Selection, a new feature introduced by our framework. The self-generated CoT data in R3V contributes to improving performance in more challenging scenarios. Test-time selection is also capable of generalizing to OOD settings.}
\label{tab:ood}
\end{table}

\begin{table*}[ht]
\centering
\footnotesize
\renewcommand\arraystretch{1.1}
\resizebox{0.95\linewidth}{!}{
\begin{tabular}{l|ccc|c|c|c|c}
    \toprule
    \multicolumn{1}{l|}{}  & \multicolumn{3}{c|}{\textbf{Logical and Numerical reasoning}}           & \multicolumn{1}{c|}{\textbf{Agentic}} & \multicolumn{1}{c|}{\textbf{Geometry}} & \multicolumn{1}{c|}{\textbf{Multi-Domain}} & \\
    \multicolumn{1}{l|}{\multirow{-2}{*}{\textbf{Methods}}}  & \textbf{TabMWP} & \textbf{ChartQA} & \multicolumn{1}{c|}{\textbf{CLEVR-Math}} & \multicolumn{1}{c|}{\textbf{MiniWob}} & \multicolumn{1}{c|}{\textbf{GeoQA}}    & \multicolumn{1}{c|}{\textbf{M\textsuperscript{3}CoT}}  & \multirow{-2}{*}{\textbf{Avg}} \\ 
    \midrule
    \rrrv &\textbf{83.27} &\textbf{57.36} &\textbf{68.81} &\textbf{82.89} &\textbf{39.25} &\textbf{54.66} &\textbf{64.37}  \\
    \quad w/o self-refine &80.87 &56.32 &64.51 &80.67 &38.33 &54.31 &62.50  \\
    \quad w/o self-select &79.72 &55.36 &64.00 &79.11 &35.81 &50.69 &60.78  \\
    \quad w/o iteration &78.53 &54.72 &64.56 &76.87 &36.07 &53.11 &60.64  \\
    \bottomrule
\end{tabular}
}
\caption{Ablation study of key components. \textit{w/o iteration} refers to the ablation of iterative training, where we sample $num\_sample\_per\_iter * num\_iter$ samples in a single pass.}
\label{tab:ablation}
\end{table*}

\begin{figure*}[t]
\centering
\includegraphics[scale=0.34]{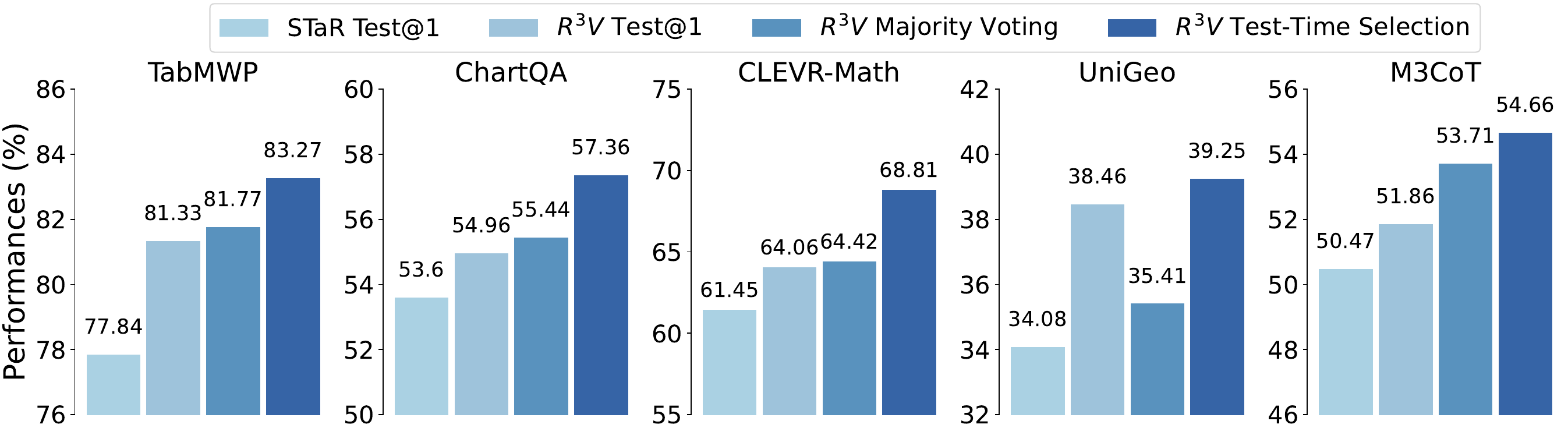}
\caption{Performance comparison of different test-time methods. Our test-time selection is robust and effective, consistently outperforming Test@1 and majority voting.}
\label{fig:test_time}
\vspace{-1.0em}
\end{figure*}

Beyond the success of the \rrrv framework on in-domain benchmarks, we are curious whether its reasoning improvements can generalize to out-of-distribution (OOD) and more difficult vision-language tasks.
To this end, we aggregated the CoT rationales self-generated by \rrrv across in-domain benchmarks and constructed positive and negative pairs for continual training on Qwen-VL.
For a fair comparison, we also included a baseline that uses only GPT-distilled positive CoT annotations.
We conducted evaluations on three challenging benchmarks: (1) \textbf{MMMU} \cite{yue2024mmmu}, a multi-discipline dataset designed to evaluate various aspects of multimodal reasoning; (2) \textbf{MathVista} \cite{lu2023mathvista}, which focuses specifically on mathematical reasoning in multimodal contexts; (3) \textbf{VCR} \cite{zellers2019recognition}, a cognition-level visual understanding benchmark that requires reasoning based on common sense and visual content.

\paragraph{\rrrv also strengthens multimodal reasoning in OOD scenarios.}
As shown in \Cref{tab:ood}, after incorporating \rrrv's self-generated CoT reasoning data, Qwen-VL significantly outperforms both the zero-shot and GPT-distilled baselines. This demonstrates that the CoT annotations synthesized by our framework not only enhance MLLM in-domain reasoning but also generalize to OOD and more challenging vision-language tasks.

\paragraph{Test-time selection generalizes to unseen tasks.}
Somewhat surprisingly, we found that the test-time selection ability does generalize to unseen tasks. For example, on MMMU, sampling three times during inference combined with our self-select mechanism (see \Cref{sec:test_select}) led to further improvement (35.63 $\to$ 38.48).
This suggests that through our self-select training, the MLLM has learned to compare multiple reasoning paths, identify errors (e.g., recognition or calculation mistakes), and eliminate incorrect options to arrive at the correct answer.

\begin{table*}[ht]
\centering
\footnotesize
\renewcommand\arraystretch{1.1}
\resizebox{0.92\linewidth}{!}{
\begin{tabular}{l|ccc|c|c|c|c}
    \toprule
    \multicolumn{1}{l|}{}  & \multicolumn{3}{c|}{\textbf{Logical and Numerical reasoning}}           & \multicolumn{1}{c|}{\textbf{Agentic}} & \multicolumn{1}{c|}{\textbf{Geometry}} & \multicolumn{1}{c|}{\textbf{Multi-Domain}} & \\
    \multicolumn{1}{l|}{\multirow{-2}{*}{\textbf{Methods}}}  & \textbf{TabMWP} & \textbf{ChartQA} & \multicolumn{1}{c|}{\textbf{CLEVR-Math}} & \multicolumn{1}{c|}{\textbf{MiniWob}} & \multicolumn{1}{c|}{\textbf{GeoQA}}    & \multicolumn{1}{c|}{\textbf{M\textsuperscript{3}CoT}}  & \multirow{-2}{*}{\textbf{Avg}} \\ 
    \midrule
    STaR &56.67 &33.44 &73.46 &76.0 &41.25 &54.06 &55.81  \\
    STaR+DPO &57.61 &32.64 &73.27 &75.33 &44.03 &52.98 &55.90  \\
    \rrrv &\textbf{59.30} &\textbf{33.92} &\textbf{79.01} &\textbf{80.11} &\textbf{45.76} &\textbf{56.08} &\textbf{59.03}  \\
    \bottomrule
\end{tabular}
}
\caption{Comparison between \rrrv and the reinforced baseline (DPO). Due to the noisy nature of CoT in multimodal scenarios, the DPO method struggles to efficiently learn from mistakes and improve performance.}
\label{tab:dpo}
\end{table*}

\section{Analysis}

This section analyzes the key factors behind the success of the \rrrv, as well as the potential challenges of self-training in multimodal reasoning tasks.

\subsection{Ablation Studies}

\paragraph{Reflection on self-generated CoT facilitates learning from mistakes.}
To validate the effectiveness of each part of our framework, we independently ablated the self-refine and self-select losses, denoted as \textit{w/o self-refine} and \textit{w/o self-select}.
As shown in \Cref{tab:ablation}, both self-refine and self-select play a crucial role in improving performance.
This highlights the value of negative samples, while our \rrrv framework's reflection mechanism (i.e., self-refine and self-select losses) serves as an effective method for learning from mistakes.

\paragraph{Iterative training process is crucial for self-improvement.}
Next, we ablated iterative training as \textit{w/o iteration}: instead of iteratively sampling and training, we sampled a large batch at once. For example, iterative self-training samples three times per round over four rounds, while \textit{w/o iteration} samples $3 \times 4=12$ times in a single pass.
This approach is similar to Rejection Sampling Fine-tuning (RFT; \citet{yuan2023scaling}), but includes our self-refine and self-select losses.
The results in \Cref{tab:ablation} demonstrate the importance of iteratation.
Although \textit{w/o iteration} produces a large number of positive and negative samples (comparable to \rrrv by our statistics), the progressive training process yields higher-quality, more diverse samples, which boosts self-training performance.

\begin{figure}[t]
    \centering
    \includegraphics[scale=0.395]{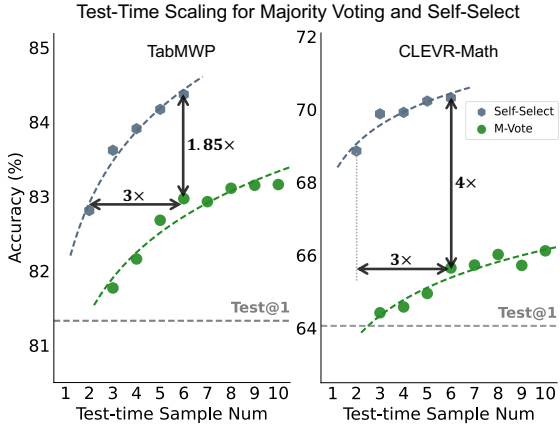}
    \caption{Comparison of scalability between test-time selection and majority voting.}
    \label{fig:scale}
    \vspace{-1.0em}
\end{figure}

\subsection{Test-time Compute}

\paragraph{Test-time self-selection boosts performance through sampling.}
One key advantage of \rrrv framework lies in its capacity to enhance performance by scaling test-time computation: during inference, we sample multiple candidate solutions and apply self-select to choose the answer.
\Cref{fig:test_time} compares self-selection with Test@1 and majority voting with a sample size of 3.
Our self-selection method consistently outperforms Test@1 and majority voting across all tasks.
While majority voting reduces noise by aggregating results, self-selection goes further by deeply comparing reasoning paths, eliminating incorrect options, and ultimately analyzing to reach the correct answer.

\paragraph{\rrrv consistently benefits from the scaling of sampling size.}
An open question is the scalability of our test-time selection. We conducted experiments with Qwen-VL on the TabMWP and CLEVR-Math benchmarks, comparing the performance of self-select and majority voting as the sample size increases.
As shown in \Cref{fig:scale}, scaling the sample size consistently improves the performance of test-time selection, achieving both higher efficiency and accuracy compared to majority voting.
Due to limitations in input length and capability of current MLLMs, performance plateaus with excessive sample size, which we believe stronger base models could address.

Our self-training framework requires no manual annotation, instead synthesizing large-scale positive and negative CoT rationales through sampling, equipping the model with the capacity for self-reflection during reasoning.
It also opens up new opportunities for boosting MLLM reasoning performance by scaling test-time computation.

\begin{figure}[t]
    \centering
    \includegraphics[scale=0.65]{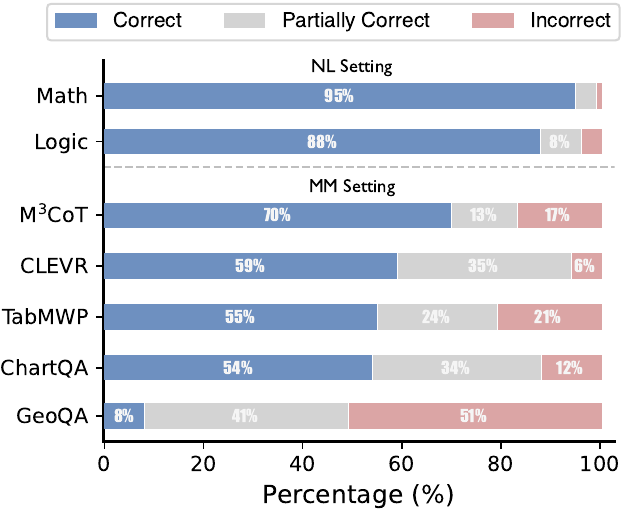}
    \caption{Proportion of correct rationales in solutions with correct answers. Multimodal CoT contains substantially more noise than text-based CoT.}
    \label{fig:mm_error}
    \vspace{-1.0em}
\end{figure}

\subsection{The Noisy Nature of Multimodal CoT}
\label{sec:noisy}

In our preliminary study, we found that the widely-used preference learning method DPO \cite{rafailov2024direct} struggles to leverage positive and negative solutions for further improvement in multimodal settings.
As shown in \Cref{tab:dpo}, equipping STaR with DPO training yields minimal improvement and falls short of our \rrrv.

To investigate DPO's failure, we closely examined the positive and negative samples self-generated by the MLLM (see details and case study in \Cref{app:error_case}).
For each task, we randomly selected 100 positive solutions based on answer correctness and manually categorized their CoT fidelity as correct, partially correct, or incorrect.
As shown in \Cref{fig:mm_error}, unlike natural language reasoning tasks (e.g., Logic, Math), multimodal CoT contains significant noise, with the proportion of fully correct CoT ranging from 8\% to 70\%.
This stems from MLLM's limited recognition capabilities, leading to flawed CoT despite correct answers, such as OCR errors.
As a result, faulty reasoning in noisy CoT is often misjudged as better solutions, making it challenging for DPO to distinguish between correct and incorrect reasoning paths and ultimately reducing performance \cite{chowdhury2024provably}.
In contrast, our reflection method avoids encouraging the generation of faulty solutions, instead guiding the model to select the correct answer through elimination, demonstrating greater efficiency in noisy multimodal CoT scenarios.

\begin{figure}[t]
    \centering
    \begin{subfigure}{\linewidth}
        \centering
        \footnotesize
        \renewcommand\arraystretch{1.1}
        \resizebox{0.8\linewidth}{!}{
            \begin{tabular}{l|c}
                \toprule
                \textbf{Methods}         & \textbf{Score}    \\
                \midrule
                Zero-shot CoT                 & 17.11           \\
                Self-Train \rrrv    & 51.72            \\
                \quad + Test-Time Selection (N=3)                     & \textbf{57.96}    \\
                \bottomrule
            \end{tabular}
        }
        \caption{The effectiveness of our self-training framework \rrrv.}
        \label{fig:geoqa_qwen2vl}
    \end{subfigure}

    \vspace{0.5em} 
    
    \begin{subfigure}{\linewidth}
        \centering
        \includegraphics[scale=0.395]{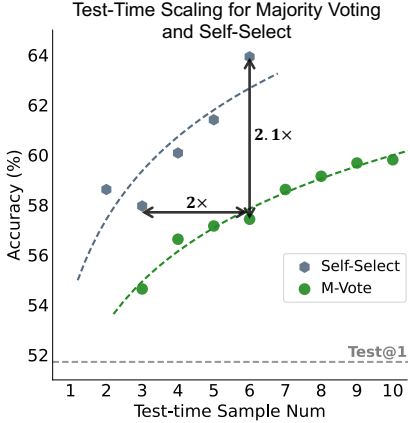}
        \caption{Scalability between test-time selection and majority voting.}
        \label{fig:scale_qwen2vl}
    \end{subfigure}
    
    \caption{Evaluation result of Qwen2-VL on GeoQA. (a) shows that our self-training approach significantly enhances performance without GPT-distilled warmup. (b) demonstrates the superior scalability of test-time selection, which boosts performance through sampling.}
    \vspace{-1.0em}
\end{figure}

\subsection{Generalization to Stronger Backbone}
To demonstrate generalizability, we applied \rrrv to the latest advanced MLLM, Qwen2-VL \cite{wang2024qwen2}
, evaluating its ability to self-improve in solving geometric problems \cite{chen2021geoqa}.
As shown in \Cref{fig:geoqa_qwen2vl}, even without GPT-distilled warmup, \rrrv achieves significant self-improvement by leveraging the model's pre-existing CoT abilities, demonstrating the \rrrv's generalizability across backbones.
More impressively, we found that test-time selection demonstrates superior scalability on Qwen2-VL, markedly outperforming majority voting, as illustrated in \Cref{fig:scale_qwen2vl}.
We hypothesize that the enhanced general capabilities of the base model further amplify the effectiveness of self-select, which we leave for future exploration.

\section{Conclusion}

The scarcity of multimodal CoT data limits the reasoning capabilities of current MLLMs.
In this paper, we take the first step toward enabling MLLMs to self-improve for better vision-language reasoning.
We propose an iterative self-training framework, \rrrv, which continuously bootstraps positive and negative solutions and improves reasoning through reflection on self-generated CoT rationales.
Meanwhile, \rrrv enables MLLMs to self-reflect on their generated solutions, offering new opportunities for boosting performance through test-time computation.
Extensive experiments and analyses demonstrate the effectiveness of our framework and the key factors behind its success.

\section*{Limitations}

As discussed in \Cref{sec:noisy}, due to the limitations of current MLLMs, the CoT annotations generated by \rrrv often contain noise. 
While our framework can self-improve performance on noisy multimodal CoT, we believe that higher-quality CoT will further enhance reasoning ability.
Due to computational constraints, our main experiments were conducted on two well-known MLLMs, LLaVA and Qwen-VL. Expanding to larger and more advanced MLLMs could yield interesting results, which we plan to explore in future work.




\newpage
\bibliography{custom}

\clearpage
\newpage
\appendix

\section{Additional Related Work}

\noindent
\textbf{Multimodal Large Language Models and Multimodal Reasoning}
Driven by the advancement of Large Language Models (LLMs), the multimodal research community has recently witnessed a domain shift from Vision-Language Models (VLMs) \cite{radford2021learning, li2022blip, cheng2023beyond} to Multimodal Large Language Models (MLLMs) \cite{achiam2023gpt, liu2023visual, chen2023shikra, cheng2024seeclick}. 
Unimodal reasoning has a strong research foundation, such as in mathematics \cite{hendrycks2021measuring} and code generation \cite{sun2024survey}. Multimodal reasoning requires models to integrate visual cues into the reasoning process \cite{zhang2023multimodal}, presenting new challenges.
Recent studies have explored synthesizing table or chart data and leveraging GPT to annotate CoT, aiming to enhance MLLM reasoning capabilities \cite{han2023chartllama, jia2024describe}. 
For instance, \citet{huang2024evochart} utilizes GPT to generate chart code and render it to obtain diverse chart reasoning samples.
In this work, we do not rely on stronger models to synthesize new reasoning samples; instead, we enable MLLMs to achieve self-improvement from self-generated CoT data.

\noindent
\textbf{Self-Training Methods}
Self-training, especially integrated with reinforcement learning from its own outputs, offers a promising avenue for model self-improvement \cite{huang2022large, gulcehre2023reinforced}.
Recent studies have applied self-training to MLLMs with the goal of enhancing image comprehension, particularly in mitigating hallucinations \cite{zhou2024calibrated, gunjal2024detecting, zhao2023beyond}.
\citet{deng2024enhancing} proposes constructing positive and negative sample pairs by perturbing images and prompts, and enhances alignment through DPO training.
In contrast, this work focuses on complex reasoning in multimodal scenarios, which requires integrating visual cues to generate step-by-step reasoning CoT.
To our knowledge, we are the first to explore self-training in the context of vision-language reasoning.

\section{Vision-Language Reasoning Benchmarks}
\label{app:bench}

\noindent
\textbf{TabMWP} \cite{lu2022dynamic}
Tabular Math Word Problems (TabMWP) is benchmark containing open-domain grade-level problems that require mathematical reasoning and calculation on table figures. 
We use the standard train/test split provided by the author.

\noindent
\textbf{ChartQA} \cite{masry2022chartqa}
We used the human-written version as the self-train benchmark, which contains more reasoning-intensive questions compared to the augmented split.
This subset contains 7,398 chart figures and question pairs, comprising both free-text and multiple-choice questions.

\noindent
\textbf{CLEVR-Math} \cite{lindstrom2022clevr}
The CLEVR-Math dataset consists of multimodal math word problems that combine text and images, where questions are posed about the state of the scene after a sequence of actions (like addition or subtraction of objects) have been applied.
We randomly sampled 10000 instances for training and used the original test set.

\begin{figure}[b]
\centering
\includegraphics[scale=0.33]{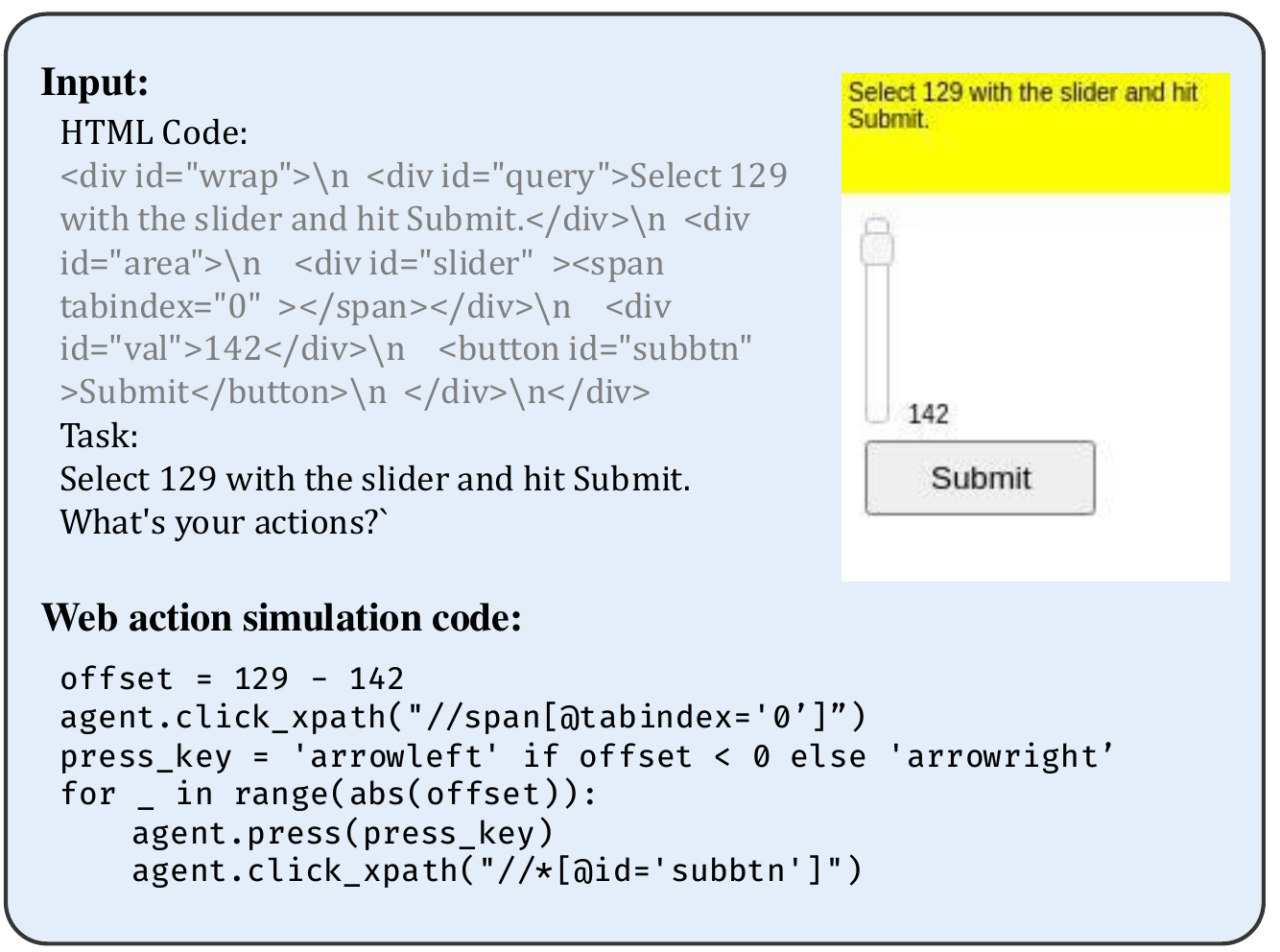}
\caption{Example for MiniWob. }
\label{fig:example_miniwob}
\vspace{-1.0em}
\end{figure}

\noindent
\textbf{MiniWob} \cite{shi2017world}
MiniWob asks MLLM to interact with a simulated Web environment.
As shown in Figure \ref{fig:example_miniwob}, the model is provided with an image of the web interface along with the html as input. 
It is then asked to generate Python code to simulate keyboard and mouse actions and complete the given task.

\noindent
\textbf{GeoQA} \cite{chen2021geoqa}
GeoQA contains 4,998 multiple-choice geometric problems from Chinese middle school exams and annotated with solving programs. 
We use human translated English version provided by UniGeo \cite{chen-etal-2022-unigeo}.

\noindent
\textbf{M$^3$CoT} \cite{chen-etal-2024-m3cot}
M$^3$CoT is a manually verified multimodal, multi-domain, multi-step visual-language reasoning dataset.
We use the official train/test splits in our \rrrv self-training process.

\paragraph{Evaluation}
For structured outputs like GPT-distilled and self-train methods, we use the benchmark's default evaluation script to calculate metrics.
For free-form outputs like the zero-shot CoT baseline, we employ GPT-4o-mini as the evaluator to assess accuracy.
For MiniWob, the simulated web environment provides an automatic reward of 0 or 1, which we use to determine the solution's correctness.

\section{Training Details}
\label{app:train_detail}


We use the Qwen-VL and LLaVA-1.5 as the base model and conducted experiments on an MLLM training infrastructure \footnote{\url{https://github.com/TideDra/VL-RLHF}}.
We show the number of training and testing samples for each dataset, along with the amount of GPT annotations in Table \ref{tab:dataset_statistics}. 
Our self-training process begins with the GPT-distilled warmup, where we fine-tune the base model using the training dataset augmented with GPT-4o CoT annotations. 
After this warm-up, the fine-tuned MLLM is employed to sample from the training dataset to build SFT, Self-Refine, and Self-Select data for our training in the \rrrv framework.
We performed self-training in either four or five iterations, depending on performance saturates.

\begin{table}[t]
\centering
\resizebox{0.9\linewidth}{!}{
\begin{tabular}{lccc}
\toprule
\textbf{Dataset} & \textbf{Train/Test} & \textbf{GPT Anno.} & \textbf{Iter\#} \\
\midrule
TabMWP    & 23059 / 7686 & 1000 &  4\\
CLEVR     & 10000 / 7955 & 1000 &  4 \\
ChartQA   & 7398 / 1250  & 800  &  5 \\
MiniWob & --    & 550   &  4   \\
M$^3$CoT     & 7973 / 2359  & 936  &  4 \\
GeoQA     & 3499 / 754  & 536  &   5 \\
\bottomrule
\end{tabular}
}
\caption{Dataset Statistics}
\label{tab:dataset_statistics} 
\end{table}

\begin{table}[t]
\centering
\resizebox{0.9\linewidth}{!}{
\begin{tabular}{lcc}
\toprule
\textbf{Hyperparameter} & \textbf{Qwen-VL} & \textbf{LLaVA-1.5} \\
\midrule
Batch Size     & \multicolumn{2}{c}{64}   \\
LR     & \multicolumn{2}{c}{3e-5}  \\
Epochs         & \multicolumn{2}{c}{3}  \\
LR Schedule    & \multicolumn{2}{c}{Constant with warmup}  \\
LR Warmup Ratio & \multicolumn{2}{c}{0.1}  \\
Weight Decay   & \multicolumn{2}{c}{0}      \\
LoRA Rank & 64 & 128 \\
LoRA alpha & 16 & 256 \\
LoRA Dropout & \multicolumn{2}{c}{0.05} \\
Optimizer      & \multicolumn{2}{c}{AdamW}  \\
\bottomrule
\end{tabular}
}
\caption{Hyperparameter Settings}
\label{tab:hyperparameters}
\end{table}

The same training hyperparameters are used across all experiments, as detailed in \Cref{tab:hyperparameters}.
We employ DeepSpeed to train MLLM using the Zero2 strategy, maintaining a global batch size of 64.

\begin{algorithm}[t]
\caption{\rrrv}\label{alg:self_refine}
\begin{algorithmic}[1]
\Require Training QA datasets $\mathcal{D}_{\text{QA}}$, subset with GPT-distilled CoT annotations $\mathcal{D}_{\text{SFT}}^{w}$, model $\mathcal{M}$, number of iterations $T$
\State Initialize $\mathcal{D}_0 = \mathcal{D}_{\text{QA}} \cup \mathcal{D}_{\text{SFT}}^{w}$ , $\mathcal{D}_{\text{pos}} = \emptyset$, $\mathcal{D}_{\text{neg}} = \emptyset$
\For{each iteration $t = 1, 2, \dots, T$}
    \State $\mathcal{D}_{\text{SFT}} = \mathcal{D}_{\text{REF}}  = \mathcal{D}_{\text{SEL}} = \emptyset$
    \State $\mathcal{M}_t \leftarrow$ SFT$(\mathcal{M}, \mathcal{D}_{t-1})$ 
    \Statex \quad \ \ \textcolor{gray}{\# sample $3$ times on training set}
    \State $\mathcal{S} \leftarrow $ sample$(\mathcal{M}_t, \mathcal{D}_{\text{QA}}, n=3)$ 
    \State $\mathcal{D}_{\text{pos}} \leftarrow \mathcal{D}_{\text{pos}} \ \cup $ eval\_pos$(\mathcal{S}, \mathcal{D}_{\text{QA}})$
    \State $\mathcal{D}_{\text{neg}} \leftarrow \mathcal{D}_{\text{neg}} \ \cup $ eval\_neg$(\mathcal{S}, \mathcal{D}_{\text{QA}})$
    \For {item index $i = 1, 2, \dots, |\mathcal{D}_{\text{QA}}|$}
        \Statex \quad \quad \quad \textcolor{gray}{\# all solutions for sample $i$} 
        \State $\mathbf{d}_+^i = \{s_1^i, s_2^i, \dots, s_k^i | s_k^i \in \mathcal{D}_{\text{pos}}^i\}$
        \State $\mathbf{d}_-^i = \{s_1^i, s_2^i, \dots, s_k^i | s_k^i \in \mathcal{D}_{\text{ neg}}^i\}$
        \Statex \textcolor{gray}{\quad \quad \quad\# the $r$-th latest pos/neg solution}
        \State $s_r^{+/-} \gets \mathbf{d}_{\text{+/-}}^i[-r]$
        \Statex \textcolor{gray}{\quad \quad \quad\# SFT}
        \If{$\exists\ s_1^+$}
            \State $\mathcal{D}_{\text{SFT}} \gets \mathcal{D}_{\text{SFT}} + s_{\text{1}}^+$ 
        \EndIf
        \Statex \textcolor{gray}{\quad \quad \quad\# self-refine}
        \If{$\exists\ (s_{\text{1}}^+, s_{\text{1}}^-)$}
            \State $\mathcal{D}_{\text{REF}} \gets \mathcal{D}_{\text{REF}} + (s_{\text{1}}^-, s_{\text{1}}^+)$
        \EndIf
        \Statex \textcolor{gray}{\quad \quad \quad\# self-select}
        \If{$\exists\ (s_{\text{1}}^+, s_{\text{1}}^-, s_{\text{2}}^-)$}
            \State $\mathcal{D}_{\text{SEL}}\gets\mathcal{D}_{\text{SEL}}+(s_{\text{1}}^+, s_{\text{1}}^-, s_{\text{2}}^-)$
        \EndIf
        \If{$\exists\ (s_{\text{1}}^+, s_{\text{2}}^+, s_{\text{1}}^-)$}
            \State $\mathcal{D}_{\text{SEL}}\gets\mathcal{D}_{\text{SEL}}+(s_{\text{1}}^+, s_{\text{2}}^+, s_{\text{1}}^-)$
        \EndIf
    \EndFor

    \State $\mathcal{D}_{t} \leftarrow \mathcal{D}_0 \cup \mathcal{D}_{\text{SFT}} \cup \mathcal{D}_{\text{REF}} \cup \mathcal{D}_{\text{SEL}}$
\EndFor

\end{algorithmic}
\end{algorithm}


\begin{figure}[t]
\centering
\vspace{-8em}
\includegraphics[scale=0.45]{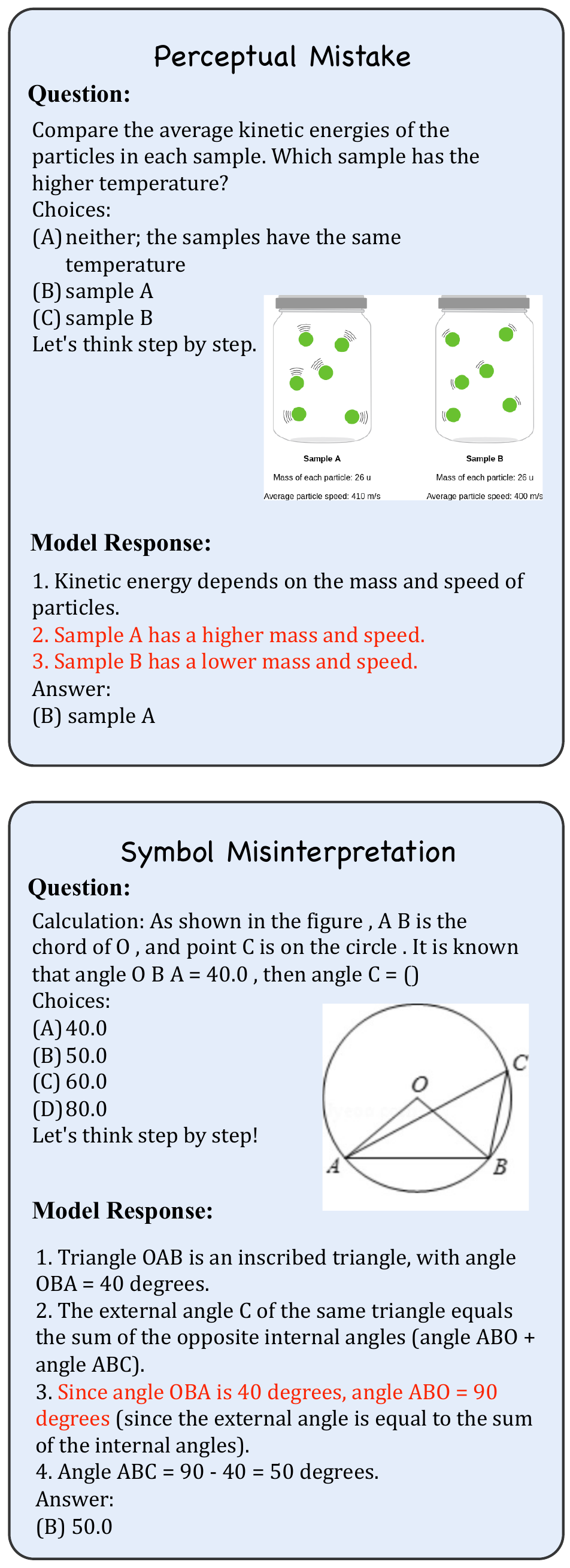}
\caption{The noisy nature of multimodal CoT, where positive solutions often contain flawed CoT steps. The perceptual error and symbol misinterpretation are marked in red.}
\label{fig:err_case}
\end{figure}

\section{Algorithms}
\label{app:alg}

Algorithm \ref{alg:self_refine} describes the overall process of \rrrv. 
The inner for-loop describes how we sample instances to build the proposed dataset, where we always select the most recent data. 
It is important to note that the sampled instances must be formatted as the data examples shown in Appendix $\ref{app:data_example}$ later.

\section{\texorpdfstring{Examples of \rrrv Multi-task Learning}{Examples of R3V Multi-task Learning}}
\label{app:data_example}

\begin{figure*}[t]
\centering
\includegraphics[scale=0.45]{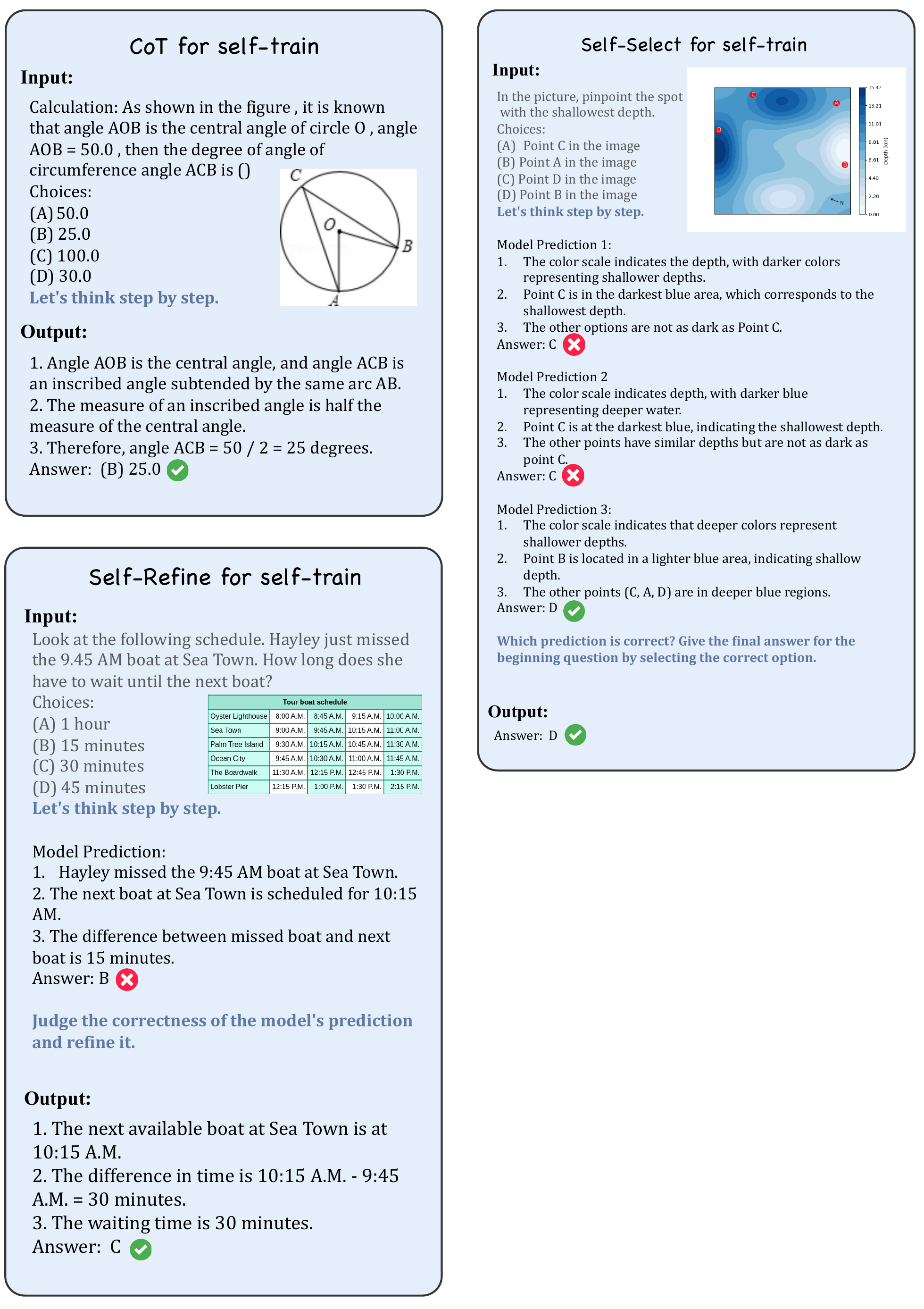}
\caption{Examples for generated data. The image displays examples of generated data for SFT, Self-Refine, and Self-Select tasks. Positive samples are marked with a green checkmark, while negative samples are marked with a red cross.}
\label{fig:data_example}
\vspace{-1.0em}
\end{figure*}

We illustrate examples of SFT, Self-Refine, and Self-Select in Figure \ref{fig:data_example}. 
The input part in each sub-figure shows the contexts such as question-choices pair and prompts used to guide the MLLM, and the output part shows the expected response from the MLLM. 
CoT prompt "Let's think step by step." will always be appended to the question-choices pairs. 
The prompts for self-refine and self-select vary slightly between multiple-choice and short-answer tasks.
Note that only the self-select prompt will be used for test-time reflection.

We add the sample generated by MLLM into both self-refine and self-select contexts using "Model Prediction" to divide with the question-choices part.
As illustrated in the figure, we use a green checkmark to indicate the positive solutions and a red cross to mark the negative ones.
It highlights that \rrrv successfully builds negative-positive rationales pairs, from which the model can learn from mistakes in negative demonstrations.
Additionally, \rrrv also builds diverse reasoning paths, ranging from completely wrong to correct rationales for the MLLM learning to choose from like human.

\section{Evolution Progress}
\label{app:evo}

\begin{figure*}[t]
    \centering
    \begin{minipage}[b]{0.48\textwidth}
        \centering
        \includegraphics[width=0.95\linewidth]{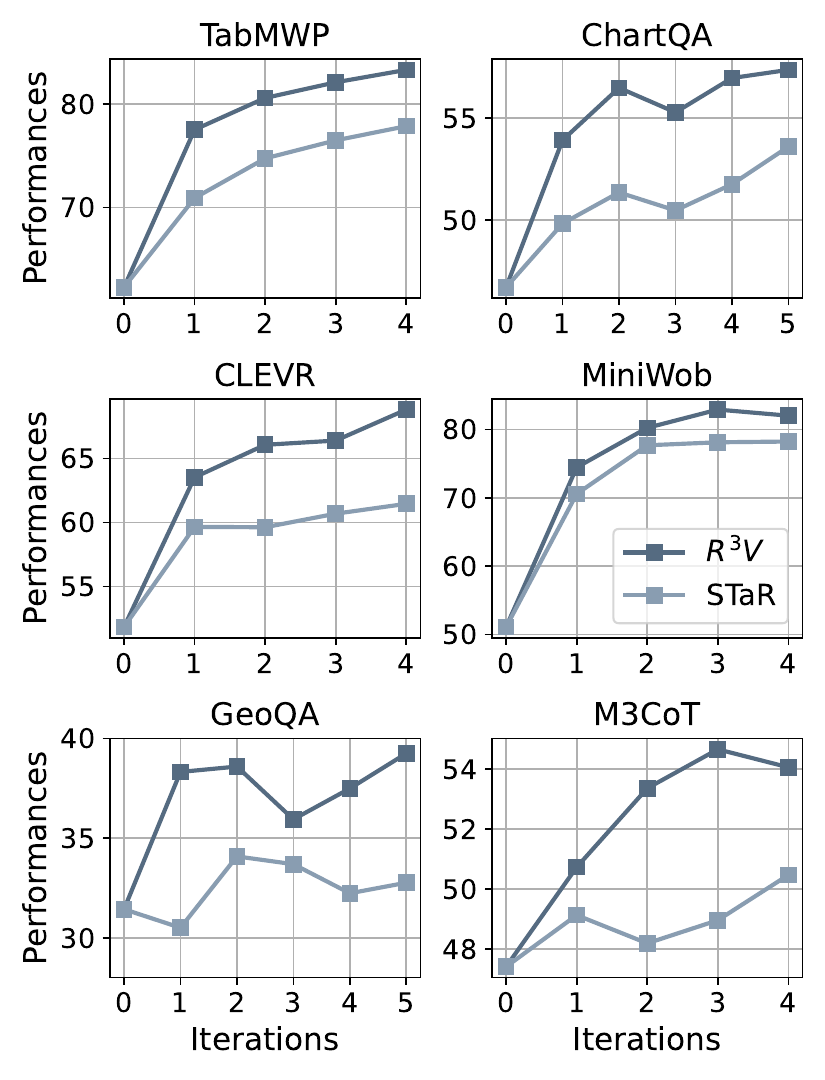}
        \subcaption{Qwen-VL}
    \end{minipage}
    \begin{minipage}[b]{0.48\textwidth}
        \centering
        \includegraphics[width=0.95\linewidth]{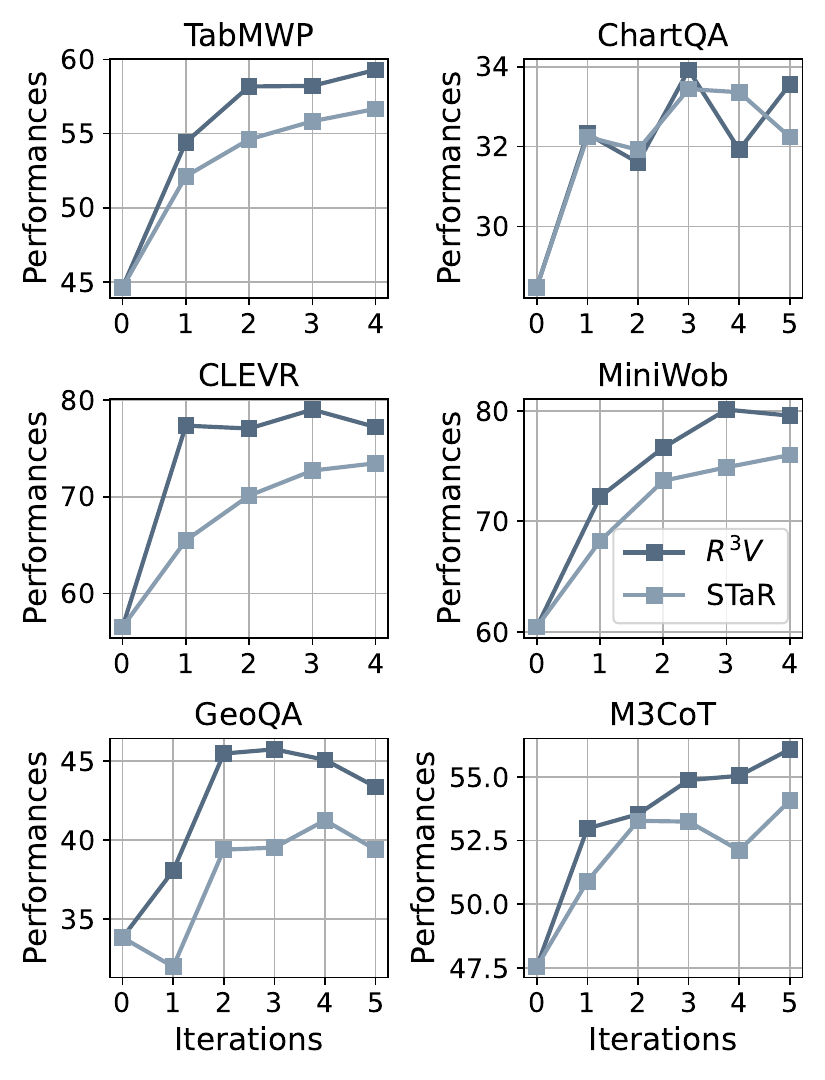}
        \subcaption{LLaVA-1.5}
    \end{minipage}
\caption{The evolution progress of \rrrv.}
\label{fig:evo}
\end{figure*}

\Cref{fig:evo} shows the evolution progress of our \rrrv framework.

\section{Noisy Nature of Multimodal CoT}
\label{app:error_case}

We manually reviewed the positive solutions generated by the Qwen-VL in our self-training process and evaluated the quality of its CoT reasoning. 
The CoT error in multimodal setting is significantly higher than samples from logical reasoning datasets \cite{liu2020logiqa, yu2020reclor} and math datasets\cite{cobbe2021training, hendrycks2021measuring} in natural language setting.
Multimodal CoT has considerable noise, such as visual perception error and symbol misinterpretation. We highlight this issue with case studies on the M$^3$CoT and GeoQA dataset in Figure \ref{fig:err_case}.



\end{document}